\documentclass[sigconf]{acmart}

\AtBeginDocument{%
  \providecommand\BibTeX{{%
    \normalfont B\kern-0.5em{\scshape i\kern-0.25em b}\kern-0.8em\TeX}}}

\setcopyright{acmcopyright}

\acmConference[ICMI '20]{ICMI20: ACM International Conference on Multimodal Interaction}{October 25--29, 2020}{Utrecht, the Netherlands}
\acmBooktitle{ICMI '20: ACM International Conference on Multimodal Interaction,
  October 25--29, 2020, Utrecht, the Netherlands}
\acmPrice{15.00}
\acmISBN{978-1-4503-XXXX-X/18/06}

\usepackage{multirow}

\usepackage{amssymb}
\usepackage{amsmath}
\usepackage{bm}
\usepackage{upgreek}
\usepackage{xcolor}

\usepackage{lipsum}

\newcommand\blfootnote[1]{%
  \begingroup
  \renewcommand\thefootnote{}\footnote{#1}%
  \addtocounter{footnote}{-1}%
  \endgroup
}

\begin{document}

\title{FairCVtest Demo: Understanding Bias in Multimodal Learning with a Testbed in Fair Automatic Recruitment}


\author{Alejandro Pe\~{n}a, Ignacio Serna, Aythami Morales and Julian Fierrez}
\affiliation{
  \institution{BiDA-Lab, Universidad Autonoma de Madrid, Spain}
 }

\renewcommand{\shortauthors}{A.Pe\~{n}a and I.Serna and A. Morales and J. Fierrez}

\begin{abstract}

 With the aim of studying how current multimodal AI algorithms based on heterogeneous sources of information are affected by sensitive elements and inner biases in the data, this demonstrator experiments over an automated recruitment testbed based on Curriculum Vitae: FairCVtest. The presence of decision-making algorithms in society is rapidly increasing nowadays, while concerns about their transparency and the possibility of these algorithms becoming new sources of discrimination are arising. This demo shows the capacity of the Artificial Intelligence (AI) behind a recruitment tool to extract sensitive information from unstructured data, and exploit it in combination to data biases in undesirable (unfair) ways. Aditionally, the demo includes a new algorithm (SensitiveNets) for discrimination-aware learning which eliminates sensitive information in our multimodal AI framework.

\end{abstract}



\begin{teaserfigure}
  \centering
  \includegraphics[width=0.85\textwidth]{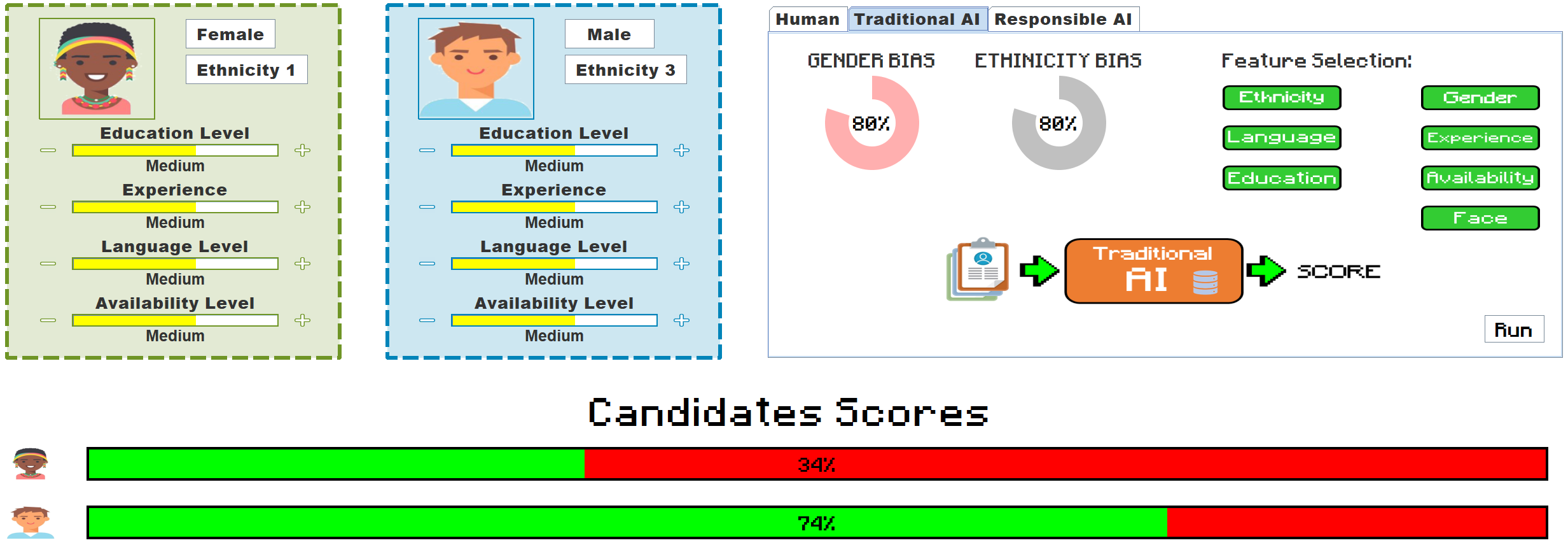}
  \caption{Capture of the interactive demonstrator with configurable parameters: candidate skills, amount of bias introduced in the groundtruth data, features used, and method for score generation: Human, Traditional AI, and Responsible AI.}
  \Description{Capture of the FairCVtest demonstrator.}
  \label{fig:teaser}
\end{teaserfigure}

\maketitle

\begin{figure*}[t]
\centering
\includegraphics[width=0.75\textwidth]{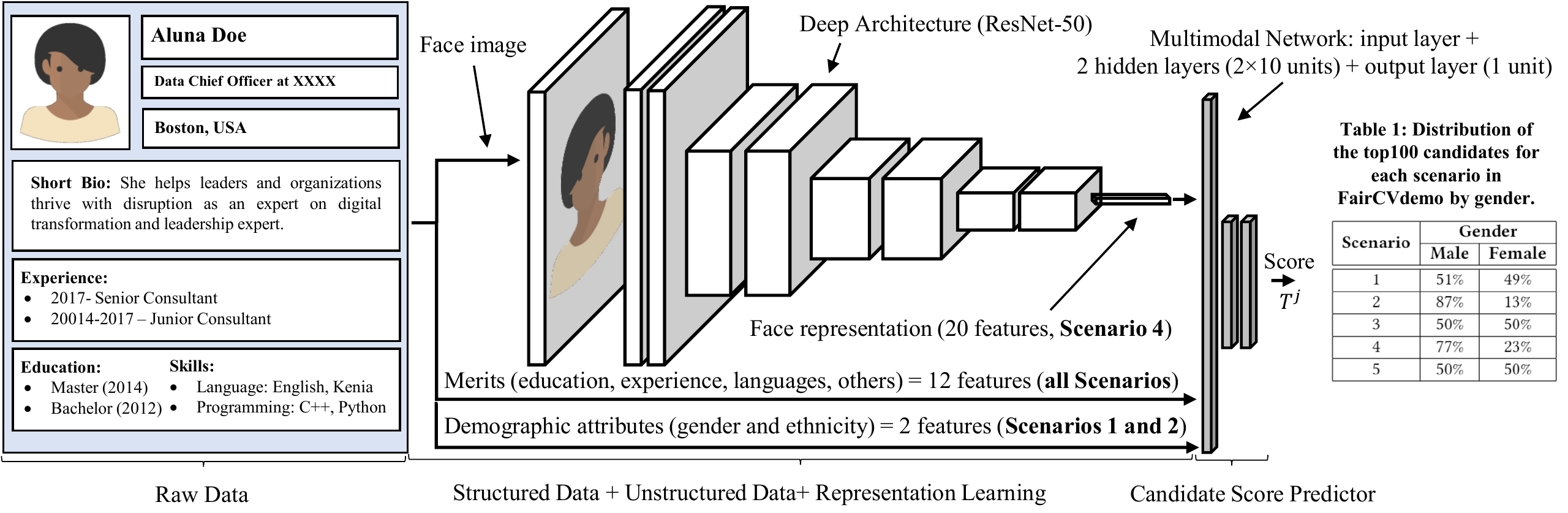} 
\caption{Multimodal learning architecture composed by a Convolutional Neural Network (ResNet-$50$) and a fully connected network used to fuse the features from different domains (image and structured data). }
\label{network}
\end{figure*}

\section{Introduction}\label{Introduction}

One of the most active areas in Artificial Intelligence (AI) is around the development of new multimodal models capable of understanding and processing information from multiple heterogeneous sources of information \cite{Baltruaitis2017MultimodalML}. Among such sources of information we can include structured data (e.g. in tables), and unstructured data from images, audio, and text. The implementation of these models in society must be accompanied by effective measures to prevent algorithms from becoming a source of discrimination. In this scenario, where multiple sources of both structured and unstructured data play a key role in algorithms' decisions, the task of detecting and preventing biases becomes even more relevant and difficult\blfootnote{This work has been supported by projects: BIBECA (RTI2018-101248-B-I00 from MINECO/FEDER), TRESPASS-ETN (H2020-MSCA-ITN-2019-860813), PRIMA (H2020-MSCA-ITN-2019-860315); and by Accenture.}. 

One application scenario of multimodal AI that is gaining in importance and where data biases can be very harmful is automated or semi-automated recruitment. The usage of AI is growing significantly in human resources departments, with video- and text-based screening software becoming increasingly common in the hiring pipeline \cite{video_interview}.

In this environment of desirable fair and trustworthy AI, the main contribution of this demo is:
\begin{itemize}
\setlength
    \item to enable interactive experimentation with a testbed for Fair Automatic Recruitment, FairCVtest\footnote{\url{https://github.com/BiDAlab/FairCVtest}}, consisting of: 1) $24$,$000$ resumes (including structured and unstructured data from face images and text), 2) various experimental protocols, and 3) popular neural network multimodal learning architectures.
\end{itemize}

This short demo paper also reports a key experiment showing the risks of traditional AI for a fair treatment of gender in automated recruitment (see Section 3) over a large population based on FairCVtest. The companion paper \cite{penna2020bias} reports additional experiments around bias effects in multimodal learning and how to overcome them.

\section{FairCVtest Demo: Description}\label{description}

The testbed visualized with this demo consists of $24$,$000$ synthetic resume profiles, as described in more detail in \cite{penna2020bias}. These resumes include $12$ features obtained from $5$ information blocks, $2$ demographic attributes (gender and ethnicity), a face photograph, and a candidate score (target score). The demo allows to analyze how multimodal algorithms are influenced by biases that are present in the target function over that testbed. Note that the testbed (FairCVtest) is available in GitHub for further research including the resumes, protocols, and baseline results on traditional and responsible AI multimodal learning methods. Fig. 1 shows a capture of the interactive demo where the user selects the following attributes for $2$ candidates: gender, ethnicity, and $4$ skill levels. Additionally, the user specifies the amount of bias in the training data and the inputs of the recruitment AI (see Fig. \ref{network}). The demo shows the candidate score for each specific combination of attributes, biases, and inputs. Finally, the user can select between three models based on Human scoring, Traditional AI, or Responsible AI trained according to the discrimination-aware method SensitiveNets \cite{SensitiveNets, blind21}.

\section{FairCVtest Demo: Experiments}\label{experiments}

In the experiments summarized here we aim at showing over a large population of the FairCVtest testbed how gender and ethnicity biases can be harmful in recruitment tools. For that we generated $5$ different versions (Scenarios, \emph{S}) of the recruitment tool with different Training (T) and additional Inputs (I). \textit{S1:} (T) Unbiased scores, (I)  Gender/Ethnicity attributes. \textit{S2:} (T) Gender-Biased scores, (I) Gender attribute. \textit{S3:} (T) Gender-Biased scores, (I) No Gender attribute. \textit{S4:} (T) Gender-Biased scores, (I) Feature embedding from the Face image. \textit{S5:} (T) Discrimination-aware training with Gender-Biased scores, (I) Feature embedding from the Face image.

Note that other scenarios can be generated interactively in the demo by changing online some parameters, see Fig. 1. In all $5$ Scenarios, we designed the candidate score predictor as a feedforward neural network with two hidden layers, both of them composed by $10$ neurons with ReLU activation, and only one neuron with sigmoid activation in the output layer, treating this task as a regression problem (see Fig. \ref{network}).

The recruitment tool was trained with the $80$\% of the synthetic profiles ($19$,$200$ CVs) described in \cite{penna2020bias}, and retaining $20$\% as validation set ($4$,$800$ CVs), each set equally distributed among gender and ethnicity, using Adam optimizer, $10$ epochs, batch size of $128$, and mean absolute error as loss metric.

We assume that the recruitment tool will be used to realize a first screening among a large list of candidates including the $4$,$800$ resumes used as validation set in our experiments. We simulate the candidates screening by choosing the top $100$ scores among them (i.e. scores with highest values). We present the distribution of these selections by gender in Table 1 (see Fig \ref{network}). In Scenarios $1$ and $3$, where the classifier shows no gender bias, we have almost no difference in the percentage of candidates selected from each gender group. On the other hand, in Scenarios $2$ and $4$ the impact of the bias is notorious, with a difference of $74\%$ in Scenario 2. However, when the sensitive features removal technique is applied \cite{SensitiveNets}, the demographic difference drops from $54\%$ to $0\%$, effectively correcting the gender bias in the training data. These results demonstrate the potential hazards of these recruitment tools in terms of fairness, and also serve to show possible ways to solve them.

\section{Conclusions}\label{conclusions}

We present FairCVtest, a testbed consisting of $24$,$000$ resumes (structured and unstructured text and face images) helpful to study how multimodal machine learning is affected by biases present in the training data. Using FairCVtest, we can study the capacity of common deep learning algorithms to expose and exploit sensitive information from commonly used structured and unstructured data. 

This demo allows to evaluate recent methods to prevent undesired effects of these biases, and to improve fairness in this AI-based recruitment framework. 


\bibliographystyle{ACM-Reference-Format}
\balance
\bibliography{acmart.bib}

\end{document}